\iftrue
\documentclass[preprint, 12pt,5p, times]{elsarticle}

\usepackage{amssymb}
\usepackage{mathtools}
\usepackage{stfloats}
\DeclarePairedDelimiter{\ceil}{\lceil}{\rceil}
\usepackage{multirow}
\usepackage{color, soul}
\usepackage{amsthm}
\usepackage{tabularx}
\usepackage{caption, setspace}
\usepackage{subcaption}
\usepackage{footnote}
\usepackage{tabularx}
\usepackage{array}
\usepackage{color, soul}
\usepackage{multirow}
\usepackage{tikz}
\newcommand*\circled[1]{\tikz[baseline=(char.base)]{
            \node[shape=circle,draw,inner sep=2pt] (char) {#1};}}
\usepackage[shortlabels]{enumitem}

\usepackage[none]{hyphenat}
\journal{Applied Energy}

\begin{document}

\title{Renewable Energy Management in Smart Home Environment via Forecast Embedded Scheduling based on Recurrent Trend Predictive Neural Network}

\author[inst1]{Mert Nak\i p\cormark[cor1]}


\ead{mnakip@iitis.pl}
\address[inst1]{Institute of Theoretical and Applied Informatics,
        Polish Academy of Sciences (PAN), 44--100 
        Gliwice,
        Poland}

\author[inst2]{Onur \c{C}opur}
\ead{onurcopur12@gmail.com}

\address[inst2]{Prime Vision, 2600 JA, Delft,Netherlands}

\author[inst3]{Emrah Biyik}
\ead{emrah.biyik@yasar.edu.tr}

\address[inst3]{Department of Energy Systems Engineering, Ya\c{s}ar University, 35100, Izmir, Turkey}

\author[inst4]{C\"{u}neyt G\"{u}zeli\c{s}}
\ead{cuneyt.guzelis@yasar.edu.tr}

\address[inst4]{Department of Electrical and Electronics Engineering,
            Ya\c{s}ar University, 35100, Izmir, Turkey}

\cortext[cor1]{Corresponding author\\ The final version of this preprint is published at Applied Energy {https://doi.org/10.1016/j.apenergy.2023.121014}.}

\begin{abstract}
Smart home energy management systems help the distribution grid operate more efficiently and reliably, and enable effective penetration of distributed renewable energy sources. These systems rely on robust forecasting, optimization, and control/scheduling algorithms that can handle the uncertain nature of demand and renewable generation. This paper proposes an advanced ML algorithm, called Recurrent Trend Predictive Neural Network based Forecast Embedded Scheduling (rTPNN-FES), to provide efficient residential demand control. rTPNN-FES is a novel neural network architecture that simultaneously forecasts renewable energy generation and schedules household appliances. By its embedded structure, rTPNN-FES eliminates the utilization of separate algorithms for forecasting and scheduling and generates a schedule that is robust against forecasting errors. This paper also evaluates the performance of the proposed algorithm for an IoT-enabled smart home. The evaluation results reveal that rTPNN-FES provides near-optimal scheduling $37.5$ times faster than the optimization while outperforming state-of-the-art forecasting techniques. 
\end{abstract}


\maketitle

\fi

\section{Introduction}

Residential loads account for a significant portion of the demand on the power system. Therefore, intelligent control and scheduling of these loads enable a more flexible, robust, and economical power system operation. Moreover, the distributed nature of the local residential load controllers increases system scalability. On the distribution level, the smart grid benefits from the increased adoption of residential demand and generation control systems, because they improve system flexibility, help to achieve a better demand-supply balance, and enable increased penetration of renewable energy sources. Increasing flexibility of the building energy demand depends on multiple developments, including accurate forecasting and effective scheduling of the loads, incorporation of renewable energy sources such as solar and wind power, and integration of suitable energy storage technologies (e.g. batteries and/or electric vehicle charging) into the building energy management system. Advanced control, optimization and forecasting approaches are necessary to operate these complex systems seamlessly. 




In this paper, in order to address this problem, we propose a novel embedded neural network architecture, called Recurrent Trend Predictive Neural Network based Forecast Embedded Scheduling (rTPNN-FES), which simultaneously forecasts the renewable energy generation and schedules the household appliances (loads). rTPNN-FES is a unique neural network architecture that enables both accurate forecasting and heuristic scheduling in a single neural network. This architecture is comprised of two main layers: 1) the Forecasting Layer which consists of replicated Recurrent Trend Predictive Neural Networks (rTPNN) with weight-sharing properties, and 2) the Scheduling Layer which contains parallel softmax layers with customized inputs each of which is assigned to a single load. In this paper, we also develop a 2-Stage Training algorithm that trains rTPNN-FES to learn the optimal scheduling along with the forecasting. {However, the proposed rTPNN-FES architecture does not depend on the particular training algorithm, and the main contributions and advantages are provided by the architectural design.} Note that the rTPNN model was originally proposed by Nakıp et al. \cite{rTPNN} for multivariate time series prediction, and its superior performance compared to other ML models was demonstrated when making predictions based on multiple time series features in the case of multi-sensor fire detection. On the other hand, rTPNN has not yet been used in an energy management system and for forecasting renewable energy generation. 

Furthermore, the advantages of using rTPNN-FES instead of a separate forecaster and scheduler are in three folds: 
\begin{enumerate}
    \item rTPNN-FES learns how to construct a schedule adapted to forecast energy generation by emulating (mimicking) optimal scheduling. Thus, the scheduling via rTPNN-FES is highly robust against forecasting errors.
    \item The requirements of rTPNN-FES for the memory space and computation time are significantly lower compared to the combination of a forecaster and an optimal scheduler.
    \item rTPNN-FES proposes a considerably high scalability for the systems in which the set of loads varies over time, e.g. adding new devices into a smart home Internet of Things (IoT) network.
\end{enumerate}
 

We numerically evaluate the performance of the proposed rTPNN-FES architecture against 7 different well-known ML algorithms combined with optimal scheduling. To this end, publicly available datasets \cite{data_PV, data_weather} are utilized for a smart home environment with 12 distinct appliances. Our results reveal that the proposed rTPNN-FES architecture achieves significantly high forecasting accuracy while generating a close-to-optimal schedule over a period of one year. It also outperforms existing techniques in both forecasting and scheduling tasks.

The remainder of this paper is organized as follows: Section~\ref{sec:RelatedWorks} reviews the differences between this paper and the state-of-the-art. Section~\ref{sec:system_design} presents the system set-up and initiates the optimization problem. Section~\ref{sec:rTPNN_FES} presents the rTPNN-FES architecture and the 2-Stage Training algorithm which is used to learn and emulate the optimal scheduling. Section~\ref{sec:Results} presents the performance evaluation and comparison. Finally, Section~\ref{sec:Conclusion} summarizes the main contributions of this paper.

\section{Related Works}\label{sec:RelatedWorks}

In this section, we present the comparison of this paper with the-state-of-the art works in three categories: 1) The works in the first category develop an optimization-based energy management system without interacting with ML. 2) The works in the second category focus on forecasting renewable energy generation using either statistical or deep learning techniques. 3) The works in the last category develop energy management systems using ML algorithms. 

\subsection{Optimization-based Energy Management Systems}
{We first review the recent works which developed optimization-based energy management systems.} In \cite{shareef2018review}, Shareef et al. gave a comprehensive summary of heuristic optimization techniques used for home energy management systems. In \cite{nezhad2022shrinking}, Nezhad et al. presented a model predictive controller for a home energy management system with loads, photovoltaic (PV) and battery electric storage. They formulated the MPC as a mixed-integer programming problem and evaluated its economic performance under different energy pricing schemes. In \cite{albogamy2022real}, Albogamy et al. utilized Lyapunov-based optimization to regulate HVAC loads in a home with battery energy storage and renewable generation. In \cite{ali2022demand}, S. Ali et al. considered heuristic optimization techniques to develop a demand response scheduler for smart homes with renewable energy sources, energy storage, and electric and thermal loads. In \cite{belli2017unified}, {G. Belli et al. resorted to mixed integer linear programming for optimal scheduling of thermal and electrical appliances in homes within a demand response framework. They utilized a cloud service provider to compute and share aggregate data in a distributed fashion. In }\cite{ali2022smart}{, variants of several heuristic optimization methods (optimal stopping rule, particle swarm optimization, and grey wolf optimization) were applied to the scheduling of home appliances under a virtual power plant framework for the distribution grid. Then, their performance was compared for three types of homes with different demand levels and profiles.}

{There is a wealth of research on optimization and model predictive controller-based scheduling of residential loads. In this literature, usually, prediction of the load demand and generation (if available) are pursued independently from the scheduling algorithm and are merely used as a constraint parameter in the optimization problem. The discrepancy in predicted and observed demand and generation may lead to poor performance and robustness issues. The proposed rTPPN-FES in this paper handles forecast and scheduling in a unified way and, therefore, provides robustness in the presence of forecasting errors. }

\subsection{Forecasting of Renewable Energy Generation}

We now briefly review the related works on forecasting renewable energy generation, which have also been reviewed in more detail in the literature, i.e. \cite{ahmed2019review, wang2019review}. 

The earlier research in this category forecast energy generation using statistical methods. For example, in \cite{kushwaha2017very}, Kushwaha et al. use the well-known seasonal autoregressive integrated moving average technique to forecast the PV generation in 20-minute intervals. In \cite{rogier2019forecasting}, Rogier et al. evaluated the performance of a nonlinear autoregressive neural network on forecasting the PV generation data collected through a LoRa-based IoT network. In \cite{fentis2019short}, Fentis et al. used Feed Forward Neural Network and Least Square Support Vector Regression with exogenous inputs to perform short-term forecasting of PV generation. In \cite{fara2021forecasting} analyzed the performances of (Autoregressive Integrated Moving Average) ARIMA and Artificial Neural Network (ANN) for forecasting the PV energy generation. In \cite{atique2019forecasting}, {Atique et al. used ARIMA with parameter selection based on Akaike information criterion and the sum of the squared estimate to forecast PV generation. In} \cite{erdem2011arma}, {Erdem and Shi analyzed the performance of autoregressive moving averages to forecast wind speed and direction in four different approaches such as decomposing the lateral and longitudinal components of the speed. In }\cite{cadenas2016wind}, {Cadenas et al. performed a comparative study between ARIMA and nonlinear autoregressive exogenous artificial neural network on the forecasting wind speed.}

The recent trend of research focuses on the development of ML and (neural network-based) deep learning techniques. In \cite{pawar2020iot}, Pawar et al. combined ANN and Support Vector Regressor (SVR) to predict renewable energy generated via PV. In \cite{corizzo2021multi}, Corizzo et al. forecast renewable energy using a regression tree with an adopted Tucker tensor decomposition. In \cite{parvez2020multi} forecast the PV generation based on the historical data of some features such as irradiance, temperature and relative humidity. In \cite{shi2017deep}, {Shi et al. proposed a pooling-based deep recurrent neural network technique to prevent overfitting for household load forecast. In }\cite{zheng2017short}{, Zheng et al. developed an adaptive neuro-fuzzy system that forecasts the generation of wind turbines in conjunction with the forecast of weather features such as wind speed.} 
In \cite{vandeventer2019short}, Vandeventer et al. used a genetic algorithm to select the parameters of SVM to forecast residential PV generation. In \cite{van2018probabilistic}, van der Meer et al. performed a probabilistic forecast of solar power using quantile regression and dynamic Gaussian process. In \cite{he2018probability}, He and Li have combined quantile regression with kernel density estimation to predict wind power density. In \cite{alessandrini2015novel}, Alessandrini et al. used an analogue ensemble method to problematically forecast wind power. In \cite{cervone2017short}{, Cervone et al. combined ANN with the analogue ensemble method to forecast the PV generations in both deterministic and probabilistic ways.} Recently in \cite{9772049}, {Guo et al. proposed a combined load forecasting method for a Multi Energy Systems (MES) based on Bi-directional Long Short-Term Memory (BiLSTM). The combined load forecasting framework is trained with a multi-tasking approach for sharing the coupling information among the loads.}

{Although there is a significantly large number of studies to forecast renewable energy generation and/or other factors related to generation, this paper differs sharply from the existing literature as it proposes an embedded neural network architecture called rTPNN-FES that performs both forecasting and scheduling simultaneously.}

\subsection{Machine Learning Enabled Energy Management Systems}

In this category, we review the recent studies that aim to develop energy management systems enabled by ML, especially for residential buildings. 

The first group of works in this category used scheduling (based on either optimization or heuristic) using the forecasts provided by an ML algorithm. In \cite{elkazaz2019optimization}, Elkazaz et al. developed a heuristic energy management algorithm for hybrid systems using an autoregressive ML for forecasting and optimization for parameter settings. In \cite{zaouali2018deep}, Zaouali et al. developed an auto-configurable middle-ware using Long-Short Term Memory (LSTM) based forecasting of renewable energy generated via PV. In \cite{shakir2020forecasting}, Shakir et al. developed a home energy management system using LSTM for forecasting and Genetic Algorithm for optimization. In \cite{manur2020smart}, Manue et al. used LSTM to forecast the load for battery utilization in a solar system in a smart home system. In \cite{ma2020hybridized} developed a hybrid system  of renewable and grid-supplied energy via exponential weighted moving average-based forecasting and a heuristic load control algorithm. In \cite{aurangzeb2022energy}, Aurangzeb et al. developed an energy management system which uses a convolutional neural network to forecast renewable energy generation. Finally, in \cite{sarker2020optimal}, in order to distribute the load and decrease the costs, Sarker et al. developed a home energy management system based on heuristic scheduling. 

The second group of works in this category developed energy management systems based on reinforcement learning. In \cite{ren2022novel}, Ren et al. developed a model-free Dueling-double deep Q-learning neural network for home energy management systems. In \cite{lissa2021deep}, Lissa et al. used ANN-based deep reinforcement learning to minimize energy consumption by adjusting the hot water temperature in the PV-enabled home energy management system. In \cite{yu2020deep}, Yu et al. developed an energy management system using a deep deterministic policy gradient algorithm. In \cite{wan2018residential}{, Wan et al. used a deep reinforcement learning algorithm to learn the energy management strategy for a residential building. In} \cite{mathew2020intelligent}{, Mathew et al. developed a reinforcement learning-based energy management system to reduce both the peak load and the electricity cost. In} \cite{liu2020optimization}{, Liu et al. developed a home energy management system using deep and double deep Q-learning techniques for scheduling home appliances.} 
In \cite{lu2021hybrid}, Lu et al. developed an energy management system with hybrid CNN-LSTM based forecasting and rolling horizon scheduling. In \cite{ji2019real}, Ji et al. developed a microgrid energy management system using the Markov decision process for modelling and ANN-based deep reinforcement learning for determining actions. 

{Deep learning-based control systems are also very popular for off-grid scenarios, as off-grid energy management systems are gaining increasing attention to provide sustainable and reliable energy services. In References }\cite{Totaro}{ and }\cite{Gao}{, the authors developed algorithms based on deep reinforcement to deal with the uncertain and stochastic nature of renewable energy sources.}

{All of these works have used ML techniques, especially deep learning and reinforcement learning, to build energy management systems. Moreover, in a recent work} \cite{nakip2021smart}{, Nak\i p et al. mimicked the scheduling via ANN and developed an energy management system using this ANN-based scheduling. However, in contrast with rTPNN-FES proposed in this paper, none of them has used ANN to generate scheduling or combined forecasting and scheduling in a single neural network architecture.}

\section{System Setup and Optimization Problem}\label{sec:system_design}
\begin{figure*}[t!]
	\centering
	\includegraphics[scale=1.2]{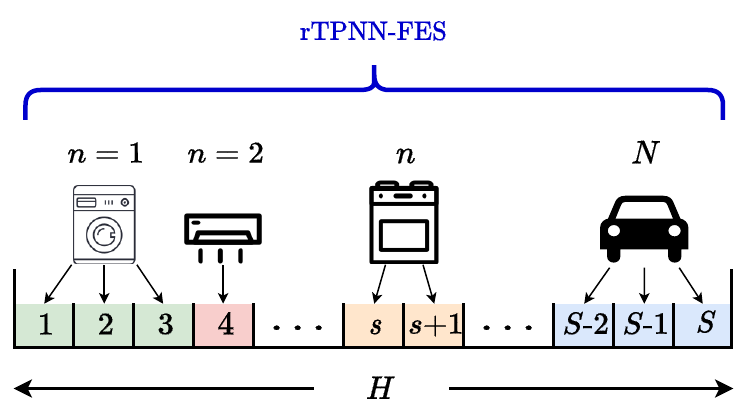}
	\caption{The illustration of the system considered by rTPNN-FES}
	\label{fig:system_design}
\end{figure*}



In this section, we present the assumptions, mathematical definitions and the optimization problem related to the system setup which is used for embedded forecasting scheduling via rTPNN-FES and shown in Figure~\ref{fig:system_design}. During this paper, rTPNN-FES is assumed to perform at the beginning of a scheduling window that consists of equal-length $S$ slots and has a total duration of $H$ in actual time (i.e. the horizon length). In addition, the length of each slot $s$ equals $H/S$, and the actual time instance at which the slot $s$ starts is denoted by $m_s$. Then, we let $g^{m_s}$ denote the power generation by the renewable energy source within slot $s$. Also, $\hat{g}^{m_s}$ denotes the forecast of $g^{m_s}$.

We let $\mathcal{N}$ be the set of devices that need to be scheduled until $H$ (in other words until the end of slot $S$), and $N$ denote the total number of devices, i.e. $|\mathcal{N}| = N$. Each device $n \in \mathcal{N}$ has a constant power consumption per slot denoted by $E_n$. In addition, $n$ should be active uninterruptedly for $a_n$ successive slots. That is, when $n$ is started, it consumes $a_n E_n$ until it stops. Moreover, we assume that the considered renewable energy system contains a battery with a capacity of $B_{max}$, where the stored energy in this battery is used via an inverter with a supply limit of $\Theta$. We assume that there is enough energy in total (the sum of the stored energy in the battery and total generation) to supply all devices within $[0, H]$.


At the beginning of the scheduling window, we forecast the renewable energy generation and schedule the devices accordingly. To this end, as the main contribution of this paper, we combine the forecaster and scheduler in a single neural network architecture, called rTPNN-FES, which shall be presented in Section~\ref{sec:rTPNN_FES}.


\textbf{Optimization Problem:} We now define the optimization problem for the non-preemptive scheduling of the starting slots of devices to minimize \emph{user dissatisfaction}. In other words, this optimization problem aims to distribute the energy consumption over slots prioritizing ``user satisfaction'', assuming that the operation of each device is uninterruptible. {In this article, we consider a completely off-grid system --which utilizes only renewable energy sources-- where it is crucial to achieve near-optimal scheduling to use limited available resources.} Recall that this optimization problem is re-solved at the beginning of each scheduling window for the available set of devices $\mathcal{N}$ using the forecast generation $\hat{g}^{m_s}$ over the scheduling window in Figure~\ref{fig:system_design}.

Moreover, for each $n \in \mathcal{N}$, there is a predefined cost of user dissatisfaction, denoted by $c_{(n, s)}$, for scheduling the start of $n$ at slot $s$. This cost can take value in the range of $[0, +\infty)$, and $c_{(n, s)}$ set to $+\infty$ if the user does not want slot $s$ to be reserved for device $n$. {As we shall explain in more detail in Section~}\ref{sec:Results}{, we determine the user dissatisfaction cost $c_{(n, s)}$ as the increasing function of the distance between $s$ and the desired start time of the considered device $n$. We should note that the definition of the user dissatisfaction cost only affects the numerical results since the proposed rTPNN-FES methodology does not depend on its definition.} 

Then, we let $x_{(n, s)}$ denote a binary schedule for the start of the activity of device $n$ at slot $s$. That is, $x_{(n, s)} = 1$ if device $n$ is scheduled to start at the beginning of slot $s$, and $x_{(n, s)} = 0$ otherwise. In addition, in our optimization program, we let $x^*_{(n, s)}$ be a binary decision variable and denote the optimal value of $x_{(n, s)}$. 
Accordingly, we define the optimization problem as follows:
\begin{equation}\label{objective_function}
min ~~ \sum_{n \in \mathcal{N}}\sum_{s=1}^{S}x^*_{(n, s)} c_{(n, s)}
\end{equation}
\noindent \textrm{subject to}
\begin{align}
    &\sum_{s=1}^{S-(a_n-1)}x^*_{(n, s)} = 1, \qquad \forall n \in \mathcal{N} \label{const:operation}\\
    &\sum_{n \in \mathcal{N}}\sum_{s'=[s-(a_n-1)]^+}^{s}E_n x^*_{(n, s')}  \leq \Theta,\quad \forall s \in \{1, \dots, S\} \label{const:inverter}\\
    &\sum_{n \in \mathcal{N}_i}\sum_{s'=[s-(a_n-1)]^+}^{s}E_n x^*_{(n, s')} \leq \hat{g}^{m_s} + B_{max},\label{const:max_battery}\\ &\qquad \qquad \qquad \qquad \qquad \qquad \forall s \in \{1,\dots, S\}\nonumber\\
    &\sum_{n \in \mathcal{N}}\sum_{s'=1}^{s}\sum_{s''=[s'-(a_n-1)]^+}^{s'}E_n x^*_{(n, s'')} \leq B + \sum_{s'=1}^{s}\hat{g}^{m_{s'}},\label{const:total_consumtion_until_now}\\ &\qquad \qquad \qquad \qquad \qquad \qquad \forall s \in \{1,\dots, S\}\nonumber
\end{align}
where $[\Xi]^+ = \Xi$ if $\Xi \geq 1$; otherwise, $[\Xi]^+ = 1$. 
The objective function (\ref{objective_function}) minimizes the total user dissatisfaction cost over all devices as ($\sum_{n \in \mathcal{N}}\sum_{s=1}^{S}x^*_{(n, s)} c_{(n, s)}$). While minimizing user dissatisfaction, the optimization problem also considers the following constraints: 
\begin{itemize}
    \item \textbf{Uniqueness and Operation} constraint in (\ref{const:operation}) ensures that each device $n$ is scheduled to start exactly at a single slot between $1$-st and $[S-(a_n-1)]$-th slot. The upper limit for the starting of the operation of device $n$ is set to $[S-(a_n-1)]$ because $n$ must operate for successive $a_n$ slots before the end of the last slot $S$.
    \item \textbf{Inverter Limitation} constraint in (\ref{const:inverter}) limits total power consumption at each slot $s$ to the maximum power of $\Theta$ that can be provided by the inverter. Note that the term $\sum_{s'=s-(a_n-1)}^{s}x^{*}_{(n,s')}$ is a convolution which equals $1$ if device $n$ is scheduled to be active at slot $s$ (i.e. $n$ is scheduled to start between $s-(a_n-1)$ and $s$). 
    \item \textbf{Maximum Storage} constraint in (\ref{const:max_battery}) ensures that the scheduled consumption at each slot $s$ does not exceed the sum of the predicted generation ($\hat{g}^{m_s}$) at this slot and the maximum energy  ($B_{max}$) that can be stored in the battery.
    \item \textbf{Total Consumption} constraint in (\ref{const:total_consumtion_until_now}) ensures that the scheduled total power consumption until each slot $s$ is not greater than the summation of the stored energy, $B$, at the beginning of the scheduling window and the total generation until $s$. {This constraint is used as we are considering a completely off-grid system.}
\end{itemize}


\section{{Recurrent Trend Predictive Neural Network based Forecast Embedded Scheduling (rTPNN-FES)}}\label{sec:rTPNN_FES}

\begin{figure*}[t!]
	\centering
	\hspace{-3cm}\includegraphics[width=20cm, height=17cm ]{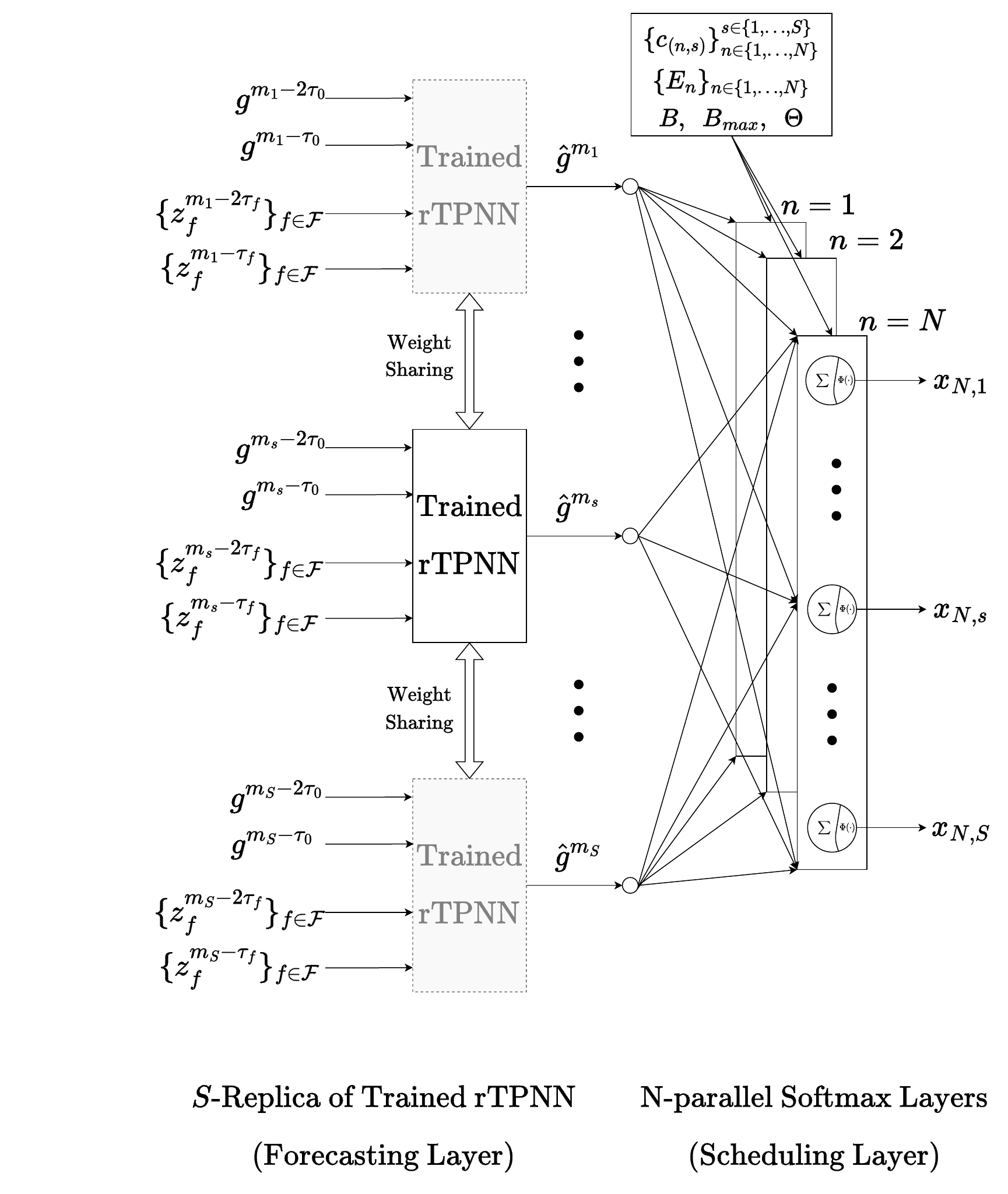}
	\caption{{Recurrent Trend Predictive Neural Network based Forecast Embedded Scheduling (rTPNN-FES)}}
	\label{fig:rTPNN-FES}
\end{figure*}

In this section, we present our rTPNN-FES neural network architecture. Figure~\ref{fig:rTPNN-FES} displays the architectural design of rTPNN-FES which aims to generate scheduling for the considered window while forecasting the power generation through this window automatically and simultaneously. To this end, rTPNN-FES is comprised of two main layers of ``Forecasting Layer'' and ``Scheduling Layer'', and it is trained using the ``2-Stage Training Procedure''. 

We let $\mathcal{F}$ be the set of features and $\mathcal{F} \equiv \{1, \dots, F\}$. In addition, $z_f^{m_s}$ denotes the value of input feature $f$ in slot $s$ which starts at $m_s$, where this feature can be considered as any external data
, such as weather predictions, that are directly or indirectly related to power generation $g^{m_s}$. We also let $\tau_f$ be a duration of time when the system developer has observed that the feature $f$ has periodicity; $\tau_0$ represents the periodicity duration for $g^{m_s}$. Note that we do not assume that the features will have a periodic nature. If there is no observed periodicity, $\tau_f$ can be set to $H$. 

As shown in {Figure~}\ref{fig:rTPNN-FES}, the inputs of rTPNN-FES are $\{g^{m_s - 2\tau_0}, g^{m_s - \tau_0}\}$ and $\{z^{m_s - 2\tau_f}_f, z^{m_s - \tau_f}_f\}$ for $f \in \mathcal{F}$, and the output of that is $\{x_{n, s}\}_{n\in\{1,\dots,N\}}^{s\in\{1,\dots,S\}}$. 

\subsection{{Forecasting Layer}}

Forecasting Layer is responsible for forecasting the power generation within the architecture of rTPNN-FES. For each slot $s$ in the scheduling window, rTPNN-FES forecasts the renewable energy generation $\hat{g}^{m_s}$ based on the collection of the past feature values for two periods, $\{z^{m_s - 2\tau_f}_f, z^{m_s - \tau_f}_f\}_{f \in \mathcal{F}}$, as well as the past generation for two periods $\{g^{m_s - 2\tau_0}, g^{m_s - \tau_0}\}$. To this end, this layer consists of $S$ parallel rTPNN models that share the same parameter set (connection weights and biases). That is, in this layer, there are $S$ replicas of a single trained rTPNN{; in other words, one may say that a single rTPNN is used with different inputs to forecast the traffic generation for each slot $s$. Therefore, all but one of the Trained rTPNN blocks are shown as transparent in Figure~}\ref{fig:rTPNN-FES}. 

The weight sharing among rTPNN models (i.e. using replicated rTPNNs) has the following advantages: 
\begin{itemize}
    \item The number of parameters in the Forecasting Layer decreases by a factor of $S$; thus reducing both time and space complexity.
    \item By avoiding rTPNN training repeated $S$ times, the training time is also reduced by a factor of $S$.
    \item Because a single rTPNN is trained on the data collected over $S$ different slots, the rTPNN can now capture recurrent trends and relationships with higher generalization ability. 
\end{itemize}

\subsubsection{Structure of rTPNN}


\begin{figure*}[h!]
	\centering
	\includegraphics[scale=0.4]{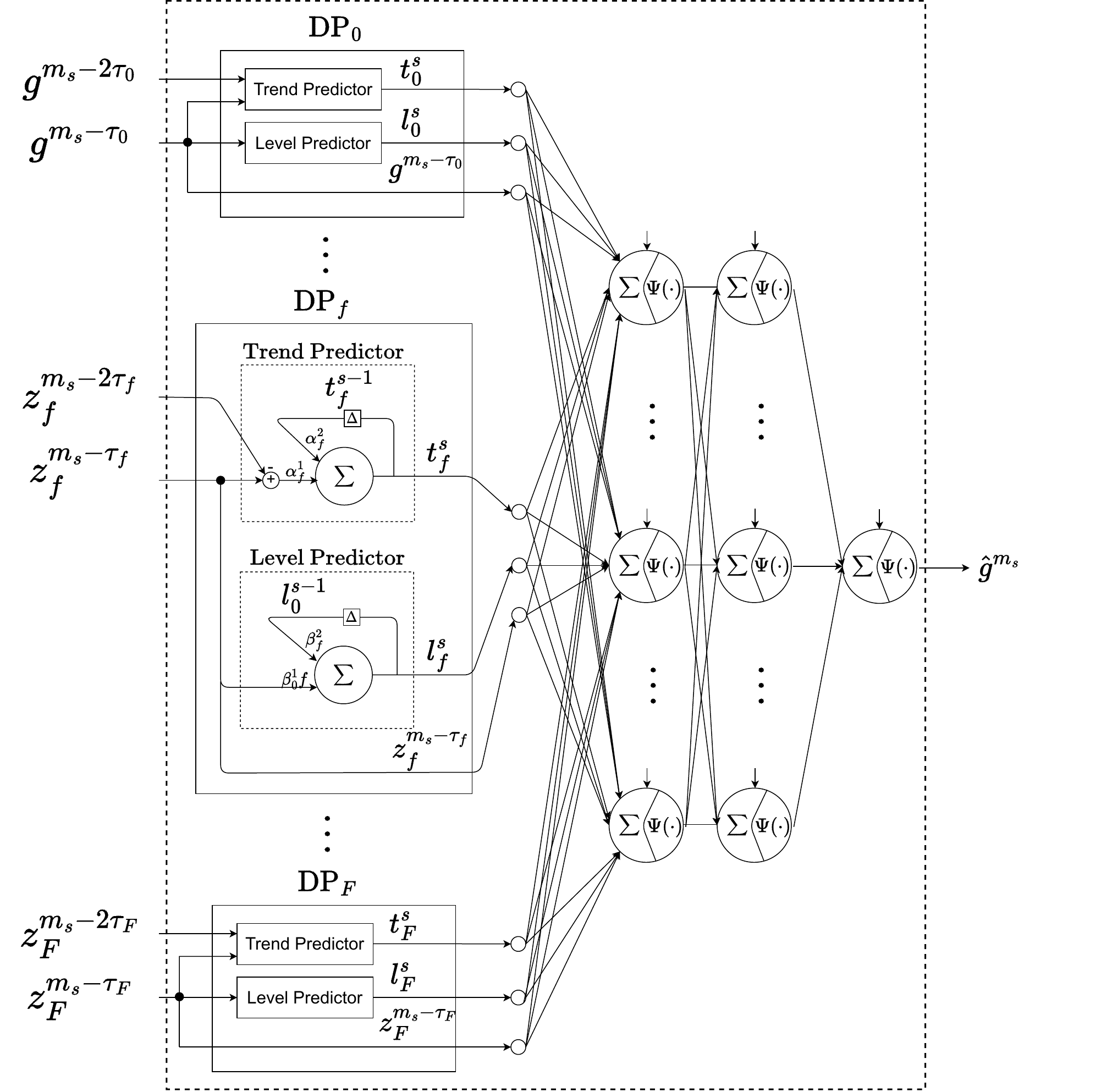}
	\caption{{The structure of rTPNN used in rTPNN-FES}}
	\label{fig:rTPNN}
\end{figure*} 

We now briefly explain the structure of rTPNN, which has been originally proposed in \cite{rTPNN}, for our rTPNN-FES neural network architecture. As shown in Figure~\ref{fig:rTPNN} displaying the structure of rTPNN, for any $s$, the inputs of rTPNN are $\{g^{m_s - 2\tau_0}, g^{m_s - \tau_0}\}$ and $\{z^{m_s - 2\tau_f}_f, z^{m_s - \tau_f}_f\}$ for $f \in \mathcal{F}$, and the output is $\hat{g}^{m_s}$. In addition, the rTPNN architecture consists of $(F+1)$ Data Processing (DP) units and $L$ fully connected layers, including the output layer.

\subsubsection{DP units}
In the architecture of rTPNN, there is one DP unit either for the past values of energy generation, denoted by $\textrm{DP}_0$ or for each time series feature $f$, denoted by $\textrm{DP}_f$. {That is, $\textrm{DP}_f$ for any feature $f$ (including $f=0$) has the same structure but its corresponding input is different for each $f$. For example, the input of $\textrm{DP}_f$ is $\{z_f^{m_s - 2\tau_f}, z_f^{m_s - \tau_f}\}$ corresponding to any time series feature $f \in \{1, \dots, F\}$ while the input of $\textrm{DP}_0$ is the past values of energy generation $\{g^{m_s - 2\tau_0}, g^{m_s - \tau_0}\}$. Thus, one may notice that $\textrm{DP}_0$ is the only unit with a special input.}

During the explanation of the DP unit, we focus on a particular instance $\textrm{DP}_f$, which is also shown in detail in Figure~\ref{fig:rTPNN}. {Using $\{z_f^{m_s - 2\tau_f}, z_f^{m_s - \tau_f}\}$ input pair, $\textrm{DP}_f$ aims to learn the relationship between this pair and each of the predicted trend $t_f^s$ and the predicted level $l_f^s$. To this end, $\textrm{DP}_f$ consists of Trend Predictor and Level Predictor sub-units each of which is a linear recurrent neuron. }

As shown in Figure~\ref{fig:rTPNN}, {Trend Predictor of $\textrm{DP}_f$ computes the weighted sum of the change in the value of feature $f$ from $m_s - 2\tau_f$ to $m_s - \tau_f$ and the previous value of the predicted trend. That is, $\textrm{DP}_f$ calculates the sum of the difference between $(z_f^{m_s - \tau_f} - z_f^{m_s - 2\tau_f})$ with connection weight of $\alpha^1_f$ and the previous value of the predicted trend $t_f^{s-1}$ with the connection weight of $\alpha^2_f$ as}

\begin{equation}\label{eq:TrendPredictor}
t_f^s=\alpha^1_f \, (z_f^{m_s - \tau_f} - z_f^{m_s - 2\tau_f})+ \alpha^2_f \, t_f^{s-1}	
\end{equation}
By calculating the trend of a feature and learning the parameters in (\ref{eq:TrendPredictor}), rTPNN is able to capture behavioural changes over time, particularly those related to the forecasting of $\hat{g}^{m_s}$.

{Level Predictor sub-unit of $\textrm{DP}_f$ predicts the level of feature value, which is the smoothed version of the value of feature $f$, using only $z_f^{m_s - \tau_f}$ and the previous state of the predicted level $l_f^{s-1}$. To this end, it computes the sum of the $z_f^{m_s - \tau_f}$ and $l_f^{s-1}$ with weights of $\beta^1_f$ and $\beta^2_f$ respectively as }
\begin{equation}\label{eq:LevelPredictor}
l_f^s=\beta^1_f \, z_f^{m_s - \tau_f}+ \beta^2_f \, l_f^{s-1}	
\end{equation}
{By predicting the level, we can reduce the effects on the forecasting of any anomalous instantaneous changes in the measurement of any other feature $f$.}

{Note that parameters $\alpha^1_f$, $\alpha^2_f$, $\beta_f^1$ and $\beta_f^2$ of Trend Predictor and Level Predictor sub-units are learned during the rTPNN training like all other parameters (i.e. connection weights).}

\subsubsection{Feed-forward of rTPNN}
We now describe the calculations performed during the execution of the rTPNN; that is, when making a prediction via rTPNN. To this end, first, let $\mathbf{W}_l$ denote the connection weight matrix for the inputs of hidden layer $l$, and $\mathbf{b}_l$ denote the vector of biases of $l$. Thus, for each $s$, the forward pass of rTPNN is as follows:

\begin{enumerate}
    \item Trend Predictors of $\textrm{DP}_0$-$\textrm{DP}_F$:
        \begin{align}
         &t_0^s=\alpha^1_0 (g^{m_s - \tau_0}-g^{m_s - 2\tau_0})+ \alpha^2_0 t_0^{s-1}, \nonumber\\
         &t_f^s=\alpha^1_f (z_f^{m_s - \tau_f}-z_f^{m_s - 2\tau_f})+ \alpha^2_f t_f^{s-1},\quad	\forall f \in \mathcal{F}
        \end{align}
        
    \item Level Predictors of $\textrm{DP}_0$-$\textrm{DP}_F$:
        \begin{align}
         & l_0^s=\beta^1_0 g^{m_s - \tau_0}+ \beta^2_0 l_0^{s-1}, \nonumber\\
         & l_f^s=\beta^1_f z_f^{m_s - \tau_0}+ \beta^2_f l_f^{s-1}, \qquad	\forall f \in \mathcal{F}
        \end{align}
        
    \item Concatenation of the outputs of $\textrm{DP}_0$-$\textrm{DP}_F$ to feed to the hidden layers:
         \begin{align}
         & \mathbf{z}^s=[t_0^s, l_0^s, g^{m_s - \tau_0}, \dots, t_F^s, l_F^s, z_F^{m_s - \tau_F}]
        \end{align}
    
    \item Hidden Layers from $l=1$ to $l=L$:
\end{enumerate}
\begin{align}
 &\mathbf{O}^s_1 = \Psi(\mathbf{W}_1 {(\mathbf{z}^s)}^T+ \mathbf{b}_1),\\
 &\mathbf{O}^s_l = \Psi(\mathbf{W}_l \mathbf{O}^s_{l-1} + \mathbf{b}_l), \quad \forall l \in \{2, \dots, L-1\}\\
 &\hat{g}^{m_s} = \Psi(\mathbf{W}_{L}\mathbf{O}^s_{L-1} + \mathbf{b}_{L}), 
\end{align}
where {${(\mathbf{z}^s)}^T$ is the transpose of the input vector $\mathbf{z}^s$}, $\mathbf{O}^s_l$ is the output vector of hidden layer $l$, and $\Psi(\cdot)$ denotes the activation function as an element-wise operator.


\subsection{Scheduling Layer}

\begin{figure}[h!]
	\centering
	\includegraphics[scale=0.48]{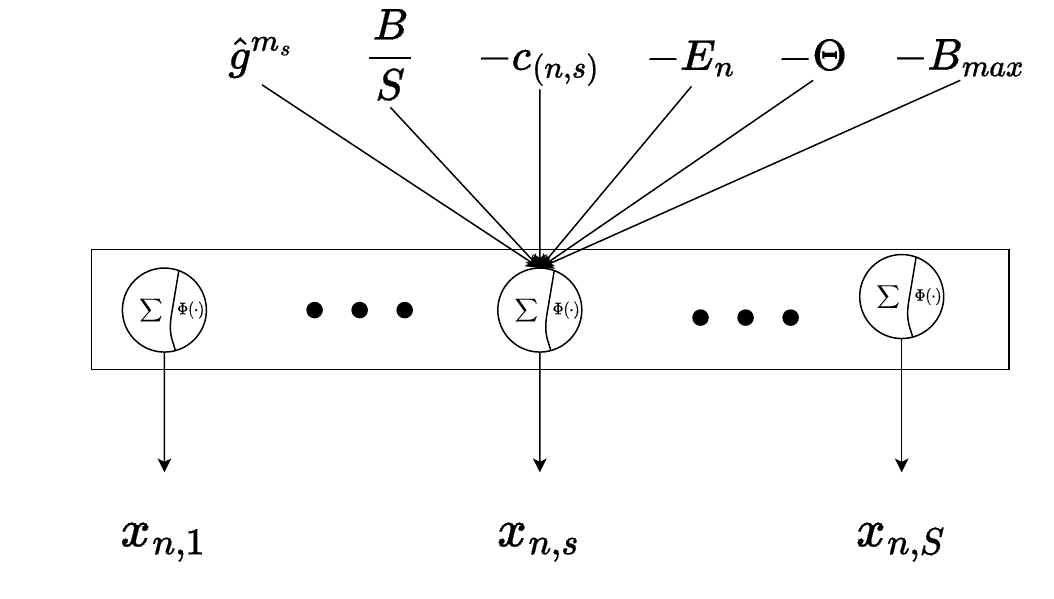}
	\caption{The structure of Scheduling Layer}
	\label{fig:scheduling_layer}
\end{figure}

The Scheduling Layer consists of $N$ parallel softmax layers, each responsible for generating a schedule for a single device's start time. A single softmax layer for device $n$ is shown in Figure~\ref{fig:scheduling_layer}. Since this layer is cascaded behind the Forecasting Layer, each device $n$ is scheduled to be started at each slot $s$ based on the output of the Forecasting Layer $\hat{g}^{m_s}$ as well as the system parameters $c_{(n,s)}$, $E_n$, $B$, $B_{max}$ and $\Theta$ for this device $n$ and this slot $s$.  

In Figure~\ref{fig:scheduling_layer}, each arrow represents a connection weight. Accordingly, for device $n$ for slot $s$ in a softmax layer of the Scheduling Layer, a neuron first calculates the weighted sum of the inputs as 
\begin{eqnarray}
        \alpha_{(n, s)} &=& w^g_{(n, s)} g^{m_s} + w^B_{(n, s)} \frac{B}{S} - w^c_{(n, s)} c_{(n, s)} \\\label{eqn:scheduling_layer}
        &&-w^{E}_{(n, s)} E_n - w^\Theta_{(n, s)} \Theta - w^{B_{max}}_{(n, s)} B_{max}\nonumber
\end{eqnarray}
{where} all connection weights of $w^g_{(n, s)}$, $w^B_{(n, s)}$, $w^c_{(n, s)}$, $w^{E}_{(n, s)}$, $w^\Theta_{(n, s)}$, and $w^{B_{max}}_{(n, s)}$ are \emph{strictly positive}. In addition, the signs of the terms are determined considering the intuitive effect of the parameter on the schedule decision for device $n$ at slot $s$. For example, the higher $g^{m_s}$ makes slot $s$ a better candidate to schedule $n$, while the higher user dissatisfaction cost $c_{(n,s)}$ makes slot $s$ a worse candidate.
In addition, a softmax activation is applied at the output of this neuron: 
\begin{equation}\label{eqn:sch_softmax}
    x_{(n, s)} ~=~ \Phi(\alpha_{(n, s)}) ~=~ \frac{e^{\alpha_{(n, s)}}}{\sum_{s=1}^{S}\alpha_{(n, s)}}
\end{equation}

\subsection{{2-Stage Training Procedure}}

We train our rTPNN-FES architecture to learn the optimal scheduling of devices as well as the forecasting of energy generation in a single neural network. To this end, we first assume that there is a collected dataset comprised of the actual values of $g^{m_s}$ and $\{z_f^{m_s}\}_{f\in\mathcal{F}}$ for $s \in \{1,\dots,S\}$ for multiple scheduling windows. {Note that rTPNN-FES does not depend on the developed 2-stage training procedure, so it can be used with any training algorithm.} For each window in this dataset, the 2-stage procedure works as follows:

\subsubsection{Stage 1 - Training of rTPNN Separately for Forecasting}
In this first stage of training, in order to create a forecaster, the rTPNN model (Figure~\ref{fig:rTPNN}) is trained separately from the rTPNN-FES architecture (Figure~\ref{fig:rTPNN-FES}). To this end, the deviation of $\hat{g}^{m_s}$ from $g^{m_s}$ for $s \in \{1, \dots,S\}$, i.e. the forecasting error of rTPNN, is measured via Mean Squared Error as
    \begin{equation}
        MSE_{\textrm{forecast}} \equiv \frac{1}{S} \sum_{s=1}^{S}(g^{m_s} - \hat{g}^{m_s})^2
    \end{equation}
We update the parameters (connection weights and biases) of rTPNN via back-propagation with gradient descent, in particular the Adam algorithm, to minimize $MSE_\textrm{forecast}$, where the initial parameters are set to parameters found in previous training. We repeat updating parameters as many epochs as required without over-fitting to the training samples. 

When Stage 1 is completed, the parameters of ``Trained rTPNN'' in Figure~\ref{fig:rTPNN-FES} are replaced by the resulting parameters found in this stage. Then, the parameters of Trained rTPNN are frozen to continue further training of rTPNN-FES in Stage 2. That is, the parameters of Trained rTPNN are not updated in Stage 2. 

\subsubsection{Stage 2 - Training of rTPNN-FES for Scheduling}
\begin{figure*}[t!]
	\centering
	\includegraphics[scale=0.4]{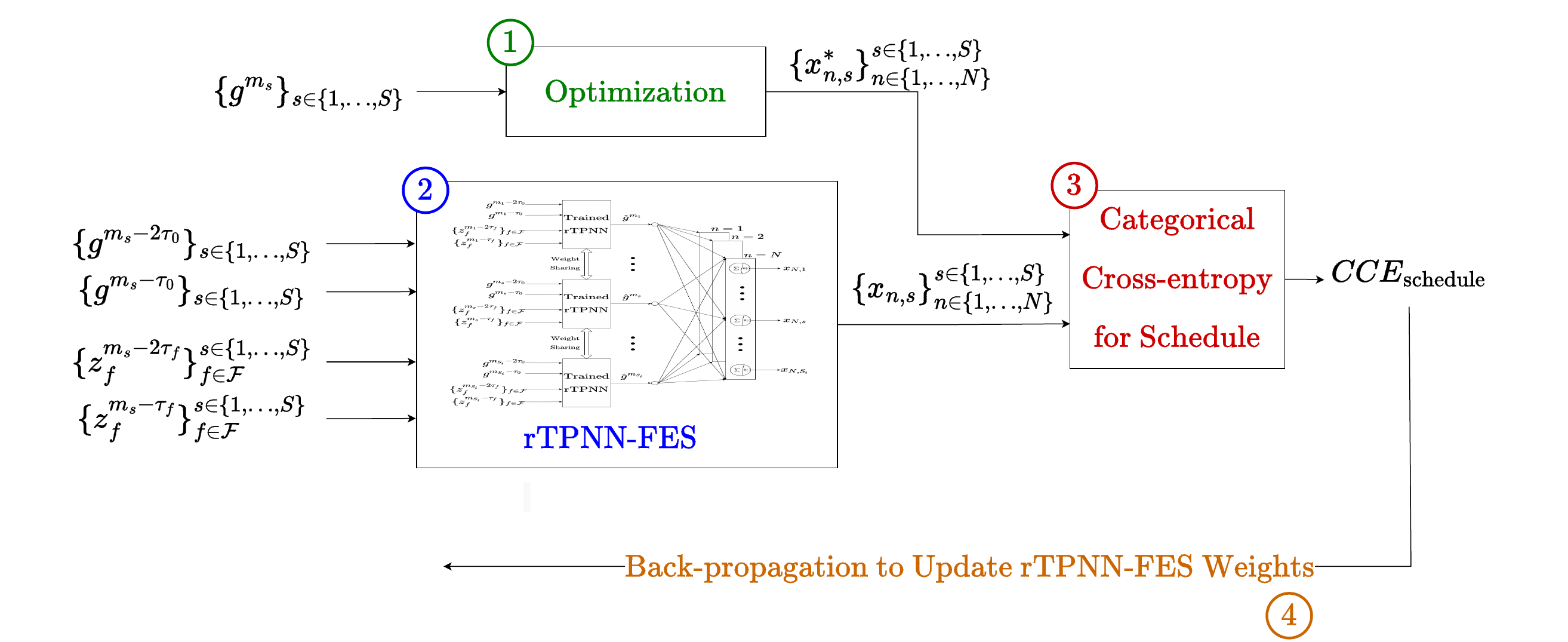}
	\caption{The steps in Stage 2 training of rTPNN-FES to learn to schedule}
	\label{fig:stage2_training}
\end{figure*}

In Stage 2 of training, in order to create a scheduler emulating optimization, the rTPNN-FES architecture (Figure~\ref{fig:rTPNN-FES}) is trained following the steps shown in Figure~\ref{fig:stage2_training}.

The steps in Stage 2 shown in Figure~\ref{fig:stage2_training} are as follows: 

\begin{enumerate}[label=\protect\circled{\arabic*}]
    \item The optimal schedule, $\{x_{n,s}^*\}_{n\in\{1,\dots,N\}}^{s\in\{1,\dots,S\}}$ is computed by solving the optimization problem given in Section~\ref{sec:system_design} in (\ref{objective_function})-(\ref{const:total_consumtion_until_now}).
    \item The feed-forward output of rTPNN-FES, $\{x_{n,s}\}_{n\in\{1,\dots,N\}}^{s\in\{1,\dots,S\}}$, which is the estimation of scheduling, is computed through (\ref{eq:TrendPredictor})-(\ref{eqn:sch_softmax}) using the architecture in Figure~\ref{fig:rTPNN-FES}. 
    \item The performance of rTPNN-FES for scheduling, i.e. total estimation error of rTPNN-FES, is measured via Categorical Cross-Entropy as
    \begin{equation}
        CCE_{\textrm{schedule}} \equiv - \sum_{n=1}^{N} \sum_{s=1}^{S} x_{n,s}^* \log(x_{n,s})
    \end{equation}
    \item The parameters (connection weights and biases) in the ``Scheduling Layers'' of rTPNN-FES are updated via back-propagation with gradient decent (using Adam optimization algorithm) to minimize $CCE_{\textrm{schedule}}$. 
\end{enumerate}
As soon as this training procedure is completed, i.e. during real-time operation, rTPNN-FES generates both forecasts of renewable energy generations, $\{\hat{g}^{m_s}\}_{s\in\{1,\dots,S\}}$ and a schedule $\{x_{n, s}\}_{{n\in\{1,\dots,N\}}}^{s\in\{1,\dots,S\}}$ that emulates the optimization.

\section{Results}\label{sec:Results}
In this section, we aim to evaluate the performance of our rTPNN-FES. To this end, during this section, we first present the considered datasets and hyper-parameter settings. We also perform a brief time-series data analysis aiming to determine the most important features for the forecasting of PV energy generation. Then, we numerically evaluate the performance of our technique and compare that with some existing techniques.  

\subsection{Methodology of Experiments}

\subsubsection{Datasets} For the performance evaluation of the proposed rTPNN-FES, we combine two publicly available datasets \cite{data_PV} and \cite{data_weather}. The first dataset \cite{data_PV} consists of hourly solar power generation (kW) of various residential buildings in Konstanz, Germany between 22-05-2015 and 12-03-2017. Within this dataset, we consider only the residential building called ``freq\_DE\_KN\_residential1\_pv'' which corresponds to 15864 samples in total. The second dataset contains weather-related information which is scraped with World Weather Online (WWO) API \cite{data_weather}. This API provides 19 features related to temperature, precipitation, illumination and wind.


\subsubsection{Experimental Set-up}

\begin{table*}[h!]
\centering
\caption{Household Appliances in the Smart Home Environment}
\begin{tabular}{|llll|}
\hline
Appliance Name  & Power Consumption (kW) & Active Duration & Desired Start Time \\\hline \hline 
Washing Machine (warm wash) & 2.3              & 2 & 14 \\
Dryer (avg. load) &  3  & 2  & 16 (earliest 15)\\
Robot Vacuum Cleaner & 0.007  & 2  & 15\\
Iron & 1.08  & 2 & 8\\
TV & 0.15  & 3 & 20\\
*Refrigerator & 0.083  & 24 & non-stop\\
Oven & 2.3  & 1 & 18\\
Dishwasher & 2  & 2 & 21\\
Electric Water Heater & 0.7  & 1 & {6, 17}\\
Central AC & 3  & 2 & {6, 18}\\
Pool Filter Pump & 1.12   & 8 & 10\\
Electric Vehicle Charger & 7.7 & 8 & 21 (earliest 18 latest 23)\\\hline    
\end{tabular}\label{table:appliances}
\end{table*}


Considering the limitations of the available dataset, we perform our experiments on a virtual residential building which is, each year, actively used between May and September. It is assumed that there are 12 different smart home appliances in active months. These appliances are shown in Table~\ref{table:appliances}, where each appliance should operate at least once a day. {Note that Electric Water Heater and Central AC operate twice a day, where the desired start times are 6:00 and 17:00 for the heater, and 6:00 and 18:00 for the AC.} In order to produce sufficient energy for the operation of these appliances, this building has its own PV system which consists of the following elements: 1) PV panels for which the generations are taken from the dataset \cite{data_PV} explained above, 2) three batteries with $13.5$ kWh capacity of each, and 3) inverter with a power rate of 10kW.

{Furthermore, during our experimental work, we set $H=24~h$, and we define the user dissatisfaction cost $c_{(n, s)}$ for each device $n$ at each slot $s$ based on the ``Desired Start Time'', which is given in Table~}\ref{table:appliances}{, as}
\begin{equation}
    c_{(n, s)} = 1 - \frac{1}{\sigma_n \sqrt{2 \pi}} \, \exp\Bigg(-\frac{1}{2}\, \bigg(\frac{s - \mu_n}{\sigma_n} \bigg)^2\Bigg)
\end{equation}
{where $\mu_n$ is the desired start time of $n$, and $\sigma_n$ is the acceptable variance for the start of $n$. The value of $\sigma_n$ is $1$ for Iron and Electric Water Heater, $2$ for TV, Oven, Dishwasher and AC, $3$ for Washing Machine and Dryer, and $5$ for Robot Vacuum Cleaner. Also, the value of $c_{(n, s)}$ is set to infinity for $s$ lower than the earliest start time and for that greater than the latest start time.}

Recall that the Water Heater and AC, which are activated twice a day, are modelled as two separate devices.

\subsubsection{Implementation and Hyper-Parameter Settings for rTPNN-FES}
We implemented rTPNN-FES by using Keras API on Python 3.7.13. The experiments are executed on the Google Colab platform with an operating system of Linux 5.4.144 and a 2.2GHz processor with 13 GB RAM. 

Forecasting Layer is trained on this platform via the adam optimizer for 40 epochs with $10^{-3}$ initial learning rate. In order to exploit the PV generation trend on daily basis, the batch size is fixed at $24$. Moreover, an $L_2$ regularization term is injected into Trend and Level Predictors in the rTPNN layer in order to avoid gradient vanishing. Finally, we used fully connected layers of rTPNN which are respectively comprised of $F+1$ and $\ceil{(F+1)/2}$ neurons with sigmoid activation. Scheduling Layer of each device is trained on the same platform also using the adam optimizer for 20 epochs with a batch size of $1$ and initial learning rate of $10^{-3}$. {Note that setting the batch size to $1$ is due to the particular implementation of rTPNN-FES which uses the Keras library.} In addition, the infinity values of $c_{(n,s)}$ are set to $100$ at the inputs of the scheduling layer in order to be able to calculate the neuron activation. We also set the periodicity $\tau_0$ of $g^{m_s}$ as $24~h$.

Furthermore, the source codes of the rTPNN-FES and experiments in this paper are shared in \cite{github_repo} in addition to the repository of the original rTPNN.

\subsubsection{{Genetic Algorithm-based Scheduling for Comparison}}\label{sec:GA}
{Genetic algorithms (GAs) have been widely used in scheduling tasks due to their ability to effectively solve complex optimization problems. GAs are able to incorporate various constraints and prior knowledge into the optimization process, making them well-suited for scheduling tasks with many constraints. GAs are also able to efficiently search through a vast search space to find near-optimal solutions, even for problems with a large number of variables }\cite{katoch2021review}{. These characteristics make GAs powerful tools for finding high-quality solutions in our experimental setup and good candidates to compare against rTPNN-FES.} 

{The experiments are executed on the Google Colab platform with the same hardware configurations of rTPNN-FES. In this experimental setting, a chromosome is a daily schedule matrix. The cross-over is made by swapping device schedules by selecting a random cross-over point out of the total number of devices and mutation is introduced by changing the scheduled time of a single device randomly with probability $0.1$. The GA application starts with sampling feasible solutions out of 5000 random solutions as an initial population. After that, 1000 new generations are simulated while the population size is fixed to 200 by making selections in an elitist style.}

\subsection{Forecasting Performance of rTPNN-FES}

We now compare the forecasting performance of rTPNN with the performances of LSTM, MLP, Linear Regression, Lasso, Ridge, ElasticNet, Random Forest as well as 1-Day Naive Forecast.\footnote{1-Day Naive Forecast equals to the original time series with 1-day lag.} {Recall that in recent literature, References} \cite{zaouali2018deep, shakir2020forecasting, manur2020smart}{ used LSTM, and Reference} \cite{fentis2019short, fara2021forecasting, pawar2020iot, lissa2021deep}{ used MLP.}

During our experimental work, the dataset is partitioned into training and test sets with the first 300 days (corresponding to 7200 samples) and the rest 361 days (corresponding to 8664 samples) respectively.

First, Table~\ref{table:forecast_performance} presents the performances of all models on both training and test sets with respect to Mean Squared Error (MSE), Mean Absolute Error (MAE), Mean Absolute Percentage Error (MAPE) and Symmetric Mean Absolute Percentage Error (SMAPE) metrics{, which are calculated as}
\begin{eqnarray}
    MSE &=& \frac{1}{S} \sum_{s=1}^{S}(g^{m_s} - \hat{g}^{m_s})^2 \label{eq:MSE_test}\\
    MAE &=& \frac{1}{S} \sum_{s=1}^{S} \Big| g^{m_s} - \hat{g}^{m_s} \Big| \label{eq:MAE}\\
    MAPE &=& \frac{100\%}{S} \sum_{s=1}^{S} \Big| \frac{g^{m_s} - \hat{g}^{m_s}}{g^{m_s}} \Big| \label{eq:MAPE}\\
    SMAPE &=& \frac{100\%}{S} \sum_{s=1}^{S}  \frac{\big|g^{m_s} - \hat{g}^{m_s}\big| }{(|g^{m_s}| + |\hat{g}^{m_s}|)/2} \label{eq:SMAPE}
\end{eqnarray}

In Table~\ref{table:forecast_performance}, the results on the test set show that rTPNN outperforms all of the other forecasters for the majority of the error metrics while some forecasters may perform better in individual metrics. However, observations on an individual error metric (without considering the other metrics) may be misleading due to its properties. For example, the MAPE of Ridge Regression is significantly low but MSE, MAE and SMAPE of that are high. The reason is that Ridge is more accurate in forecasting samples with high energy generation than forecasting those with low generations. Moreover, rTPNN is shown to have high generalization ability since it performs well for both training and test sets with regard to all metrics. Also, only rTPNN and LSTM are able to achieve better performances than the benchmark performance of the 1-Day Naive Forecast with respect to MSE, MAE and SMAPE. 

{We also see that SMAPE yields significantly larger values than those of other metrics (including MAPE) because SMAPE takes values in $[0, 200]$ and has a scaling effect as a result of the denominator in (}\ref{eq:SMAPE}{). In particular, the absolute deviation of forecast values from the actual values is divided by the sum of those. Therefore, under- and over-forecasting have different effects on SMAPE, where under-forecasting results in higher SMAPE.} 

\begin{table*}[t!]
\centering
	\renewcommand{\arraystretch}{1}
	\normalsize
\caption{Comparison of the forecasting performance of rTPNN with that of state-of-the-art forecasters with respect to MSE, MAE, MAPE, and SMAPE excluding nights}
\begin{tabular}{|c||cccc||cccc|}
\hline
\multirow{2}{*}{\textbf{Forecasting Methods}} &
  \multicolumn{4}{c||}{\textbf{Training Set}} &
  \multicolumn{4}{c|}{\textbf{Test Set}} \\ \cline{2-9} 
 &
  \multicolumn{1}{c|}{\textbf{MSE}} &
  \multicolumn{1}{c|}{\textbf{MAE}} &
  \multicolumn{1}{c|}{\textbf{MAPE}} &
  \textbf{SMAPE} &
  \multicolumn{1}{c|}{\textbf{MSE}} &
  \multicolumn{1}{c|}{\textbf{MAE}} &
  \multicolumn{1}{c|}{\textbf{MAPE}} &
  \textbf{SMAPE} \\ \hline
\textbf{rTPNN} &
  \multicolumn{1}{c|}{2.23} &
  \multicolumn{1}{c|}{1.13} &
  \multicolumn{1}{c|}{3.72} &
  51.84 &
  \multicolumn{1}{c|}{2.58} &
  \multicolumn{1}{c|}{1.21} &
  \multicolumn{1}{c|}{10.67} &
  54.42 \\ \hline
\textbf{LSTM} &
  \multicolumn{1}{c|}{2.18} &
  \multicolumn{1}{c|}{1.18} &
  \multicolumn{1}{c|}{4.95} &
  54.83 &
  \multicolumn{1}{c|}{2.56} &
  \multicolumn{1}{c|}{1.26} &
  \multicolumn{1}{c|}{13.59} &
  57.98 \\ \hline
\textbf{MLP} &
  \multicolumn{1}{c|}{2.77} &
  \multicolumn{1}{c|}{1.35} &
  \multicolumn{1}{c|}{6.33} &
  60.57 &
  \multicolumn{1}{c|}{3.09} &
  \multicolumn{1}{c|}{1.42} &
  \multicolumn{1}{c|}{14.25} &
  63.06 \\ \hline
\textbf{Linear Regression} &
  \multicolumn{1}{c|}{2.78} &
  \multicolumn{1}{c|}{1.28} &
  \multicolumn{1}{c|}{4.92} &
  57.71 &
  \multicolumn{1}{c|}{3.16} &
  \multicolumn{1}{c|}{1.35} &
  \multicolumn{1}{c|}{6.08} &
  60.38 \\ \hline
\textbf{Lasso Regression} &
  \multicolumn{1}{c|}{8.61} &
  \multicolumn{1}{c|}{2.12} &
  \multicolumn{1}{c|}{4.06} &
  88.68 &
  \multicolumn{1}{c|}{8.7} &
  \multicolumn{1}{c|}{2.14} &
  \multicolumn{1}{c|}{11.16} &
  90.68 \\ \hline
\textbf{Ridge Regression} &
  \multicolumn{1}{c|}{2.78} &
  \multicolumn{1}{c|}{1.29} &
  \multicolumn{1}{c|}{4.93} &
  57.74 &
  \multicolumn{1}{c|}{3.16} &
  \multicolumn{1}{c|}{1.36} &
  \multicolumn{1}{c|}{6.11} &
 60.41\\ \hline
\textbf{ElasticNet Regression} &
  \multicolumn{1}{c|}{8.61} &
  \multicolumn{1}{c|}{2.12} &
  \multicolumn{1}{c|}{4.06} &
  88.68 &
  \multicolumn{1}{c|}{8.7} &
  \multicolumn{1}{c|}{2.14} &
  \multicolumn{1}{c|}{11.16} &
  90.69 \\ \hline
\textbf{RandomForestRegressor} &
  \multicolumn{1}{c|}{0.3} &
  \multicolumn{1}{c|}{0.41} &
  \multicolumn{1}{c|}{1.5} &
  24.75 &
  \multicolumn{1}{c|}{3.18} &
  \multicolumn{1}{c|}{1.36} &
  \multicolumn{1}{c|}{6.82} &
  60 \\ \hline
\textbf{1-Day Naive Forecast} &
  \multicolumn{1}{c|}{3.68} &
  \multicolumn{1}{c|}{1.25} &
  \multicolumn{1}{c|}{2.76} &
  56.63 &
  \multicolumn{1}{c|}{4.25} &
  \multicolumn{1}{c|}{1.37} &
  \multicolumn{1}{c|}{1.26} &
  58.29 \\ \hline
\end{tabular}
\label{table:forecast_performance}
\end{table*}

\begin{figure}[h!]
	\centering
	\hspace*{-0.9cm}\includegraphics[scale=0.25]{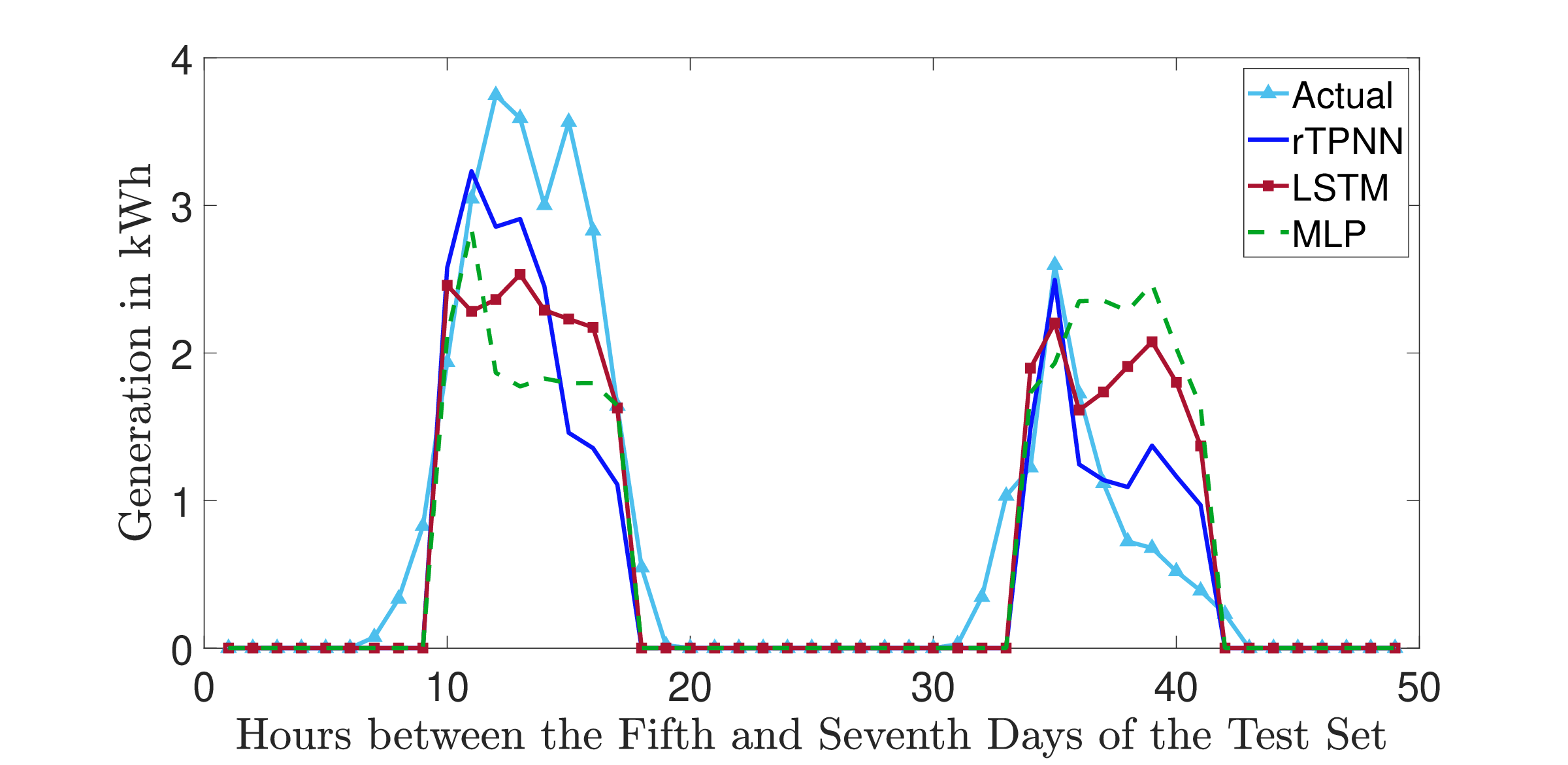}
	\caption{{Forecasting results of the three most competitive models (rTPNN, LSTM and MLP) with respect to results in Table~}\ref{table:forecast_performance}{ for the time between fifth and seventh days in the test set}}
	\label{fig:forecasting_time}
\end{figure}

Next, in Figure~\ref{fig:forecasting_time}, we present the actual energy generation between the fifth and the seventh days of the test set as well as those forecast by the best three techniques (rTPNN, LSTM and MLP). Our results show that the predictions of rTPNN are the closest to the actual generation within the predictions of these three techniques. In addition, we see that rTPNN can successfully capture both increases and decreases in energy generation while LSTM and MLP struggle to predict sharp increases and decreases.

\begin{figure}[h!]
	\centering
	\hspace*{-0.5 cm}\includegraphics[scale=0.25]{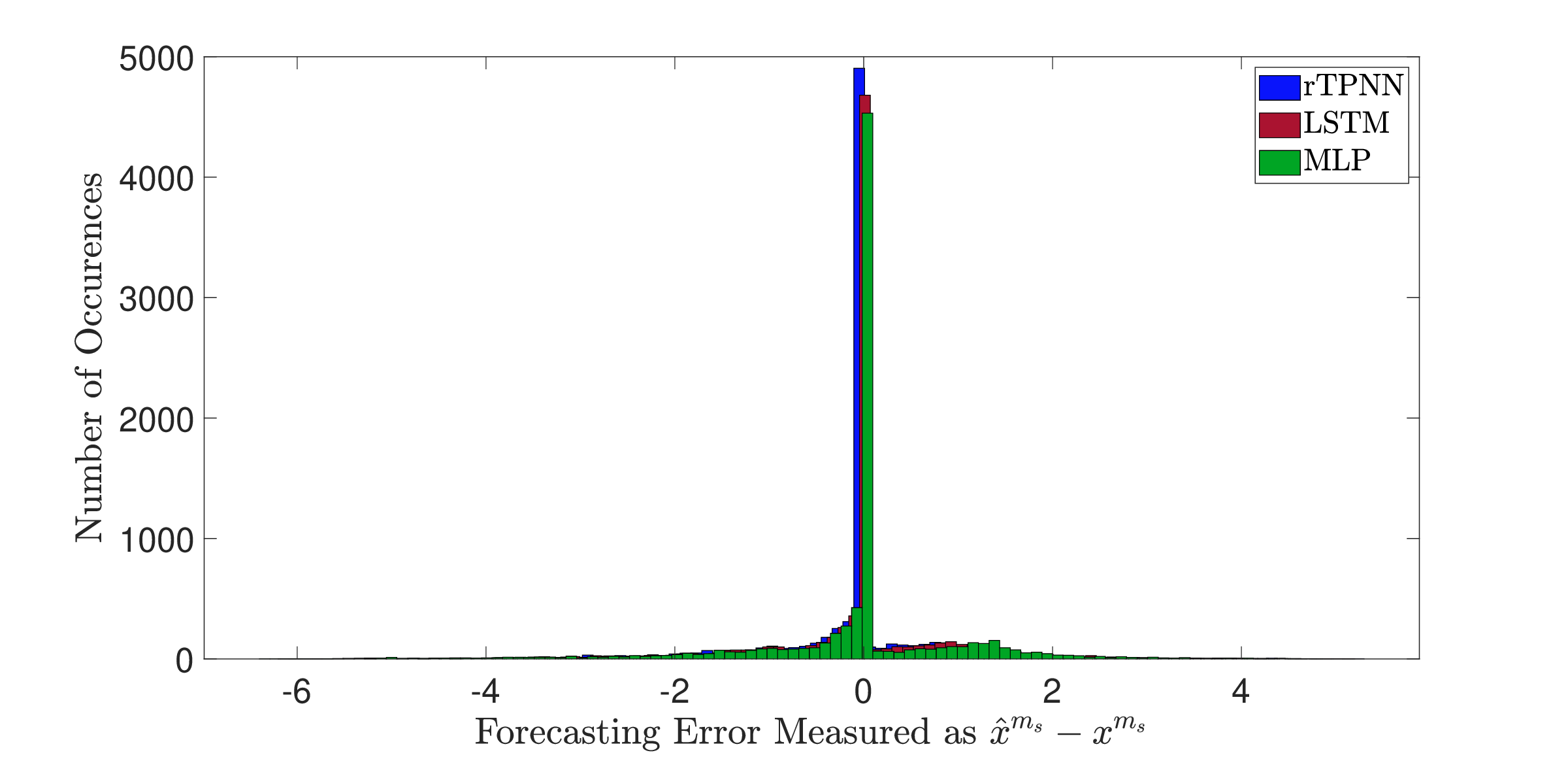}
	\caption{{Histogram of the forecasting error in kW measured as $(\hat{g}^{m_s}-g^{m_s})$ for each $m_s$ in the test set}}
	\label{fig:histogram_error}
\end{figure}

Finally, Figure~\ref{fig:histogram_error} {displays the histogram of the forecasting error that is realized by each of rTPNN, LSTM, and MLP on the test set. Our results in this figure show that the forecasting error of rTPNN is around zero for the significantly large number of samples (around 5000 out of 8664 samples). We also see that the absolute error is smaller than $2$ for $93\%$ of the samples. We also see that the overall forecasting error is lower for rTPNN than both LSTM and MLP.}

\subsection{Scheduling Performance of rTPNN-FES}

We now evaluate the scheduling performance of rTPNN-FES for the considered smart home energy management system. To this end, we compare the schedule generated by rTPNN-FES with that by optimization (solving (\ref{objective_function})-(\ref{const:total_consumtion_until_now})) using actual energy generations {as well as the GA-based scheduling (presented in Section~}\ref{sec:GA}). 
Note that although the schedule generated by the optimization using actual generations is the best achievable schedule, it is practically not available due to the lack of future information about the actual generations. 

\begin{figure*}[t!]
	\centering
	\includegraphics[scale=0.4]{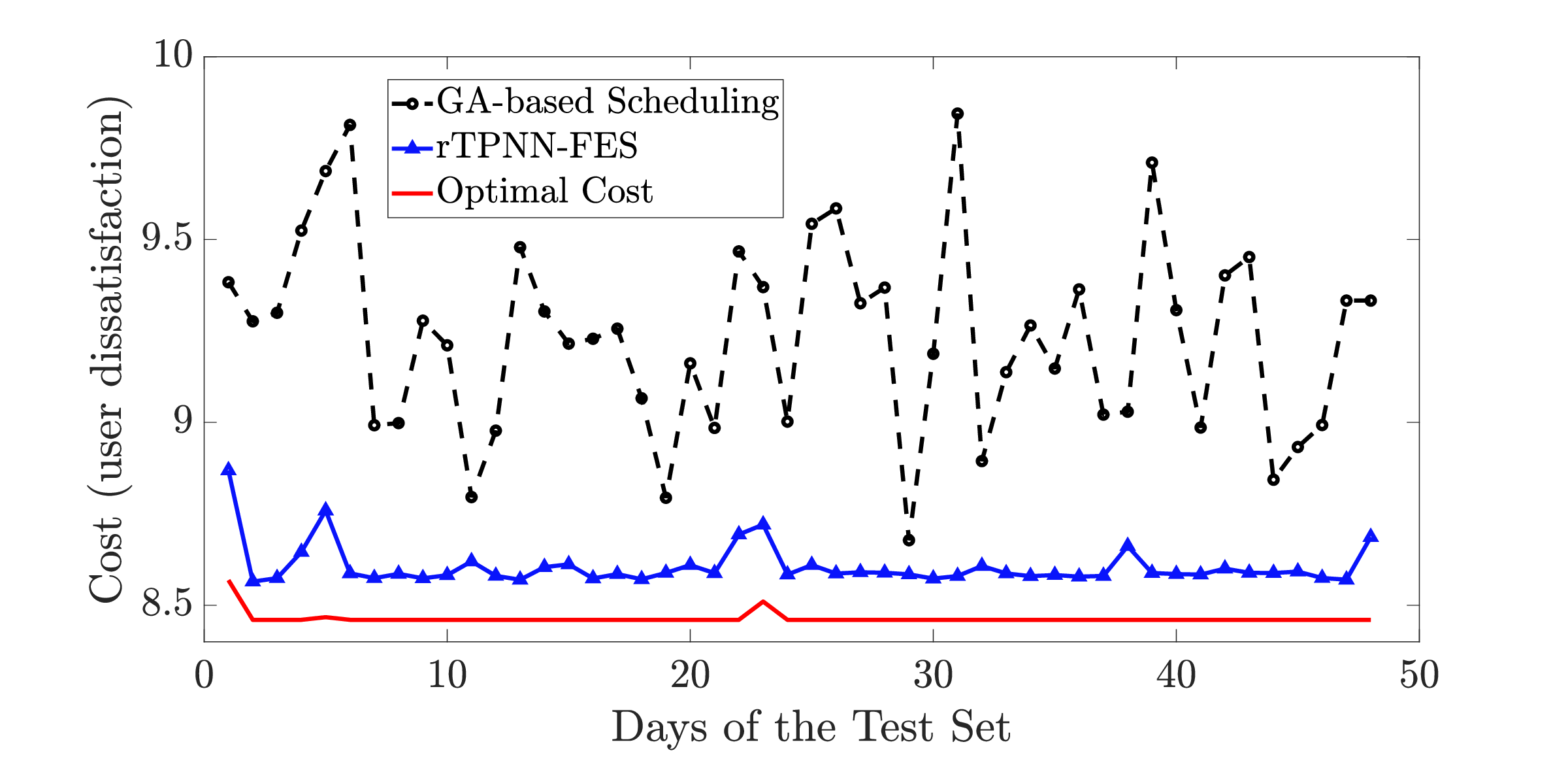}\vspace{0.5cm}
	\includegraphics[scale=0.4]{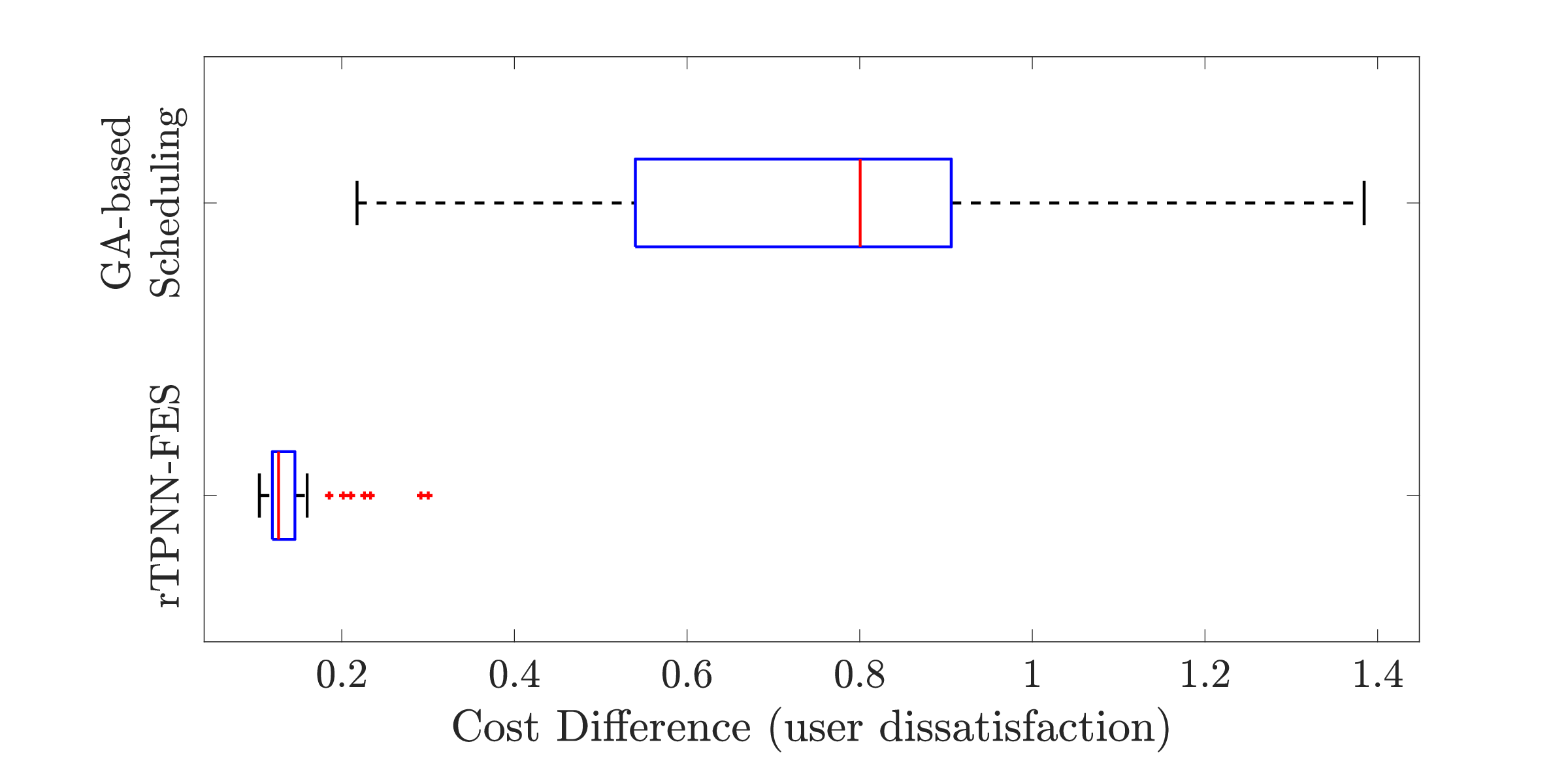}
	\caption{{Comparison of rTPNN-FES against the optimal scheduling and GA-based scheduling with respect to the scheduling cost (top) for the days of the test set and (bottom) as the boxplot of the cost difference.}}
	\label{fig:cost}
\end{figure*}

Figure~\ref{fig:cost} (top) {displays the comparison of rTPNN-FES against the optimal scheduling and the GA-based scheduling regarding the cost value for the days of the test set. In this figure, we see that rTPNN-FES significantly outperforms GA-based scheduling achieving close-to-optimal cost. In other words, the user dissatisfaction cost -- which is defined in (}\ref{objective_function}{) -- of rTPNN-FES is significantly lower than the cost of GA-based scheduling, and it is slightly higher than that of optimal scheduling. The average cost difference between rTPNN-FES and optimal scheduling is $1.3\%$ and the maximum difference is about $3.48\%$.}

{Furthermore, Figure}~\ref{fig:cost}{ (bottom) displays the summary of the statistics for the cost difference between rTPNN-FES and the optimal scheduling as well as the difference between GA-based and optimal scheduling as a boxplot.} In Figure~\ref{fig:cost} (bottom), {we first see that the cost difference is significantly lower for rTPNN-FES, where even the upper quartile of rTPNN-FES is smaller than the lower quartile of GA-based scheduling.} We also see that the median of the cost difference between rTPNN-FES and optimal scheduling is $0.13$ and the upper quartile of that is about $0.146$. That is, the cost difference is less than $0.146$ for $75 \%$ of the days in the test set. In addition, we see that there are only $7$ outlier days for which the cost is between $0.19$ and $0.3$. According to the results presented in Figure~\ref{fig:cost}, rTPNN-FES can be considered as a successful heuristic with a low increase in cost.

\subsection{Evaluation of the Computation Time}

In Table~\ref{table:forecaster_times}, we present measurements on the training and execution times of each forecasting model. Our results first show that the execution time of rTPNN ($0.17~ms$) is comparable with the execution time of LSTM and highly acceptable for real-time applications. On the other hand, the training time measurements show that the training of rTPNN takes longer than that of other forecasting models. Accordingly, one may say that there is a trade-off between training time and the forecasting performance of rTPNN.

\begin{table}[h!]
\centering
\renewcommand{\arraystretch}{1.2}
\setlength{\tabcolsep}{6pt}
\normalsize
\caption{Training and Execution Times for Forecasting}
\begin{tabular}{|c|c|c|}
\hline
\begin{tabular}{c} Forecasting \\ Methods \end{tabular} & \begin{tabular}[c]{@{}c@{}}Training Time\\  (seconds)\end{tabular} & \begin{tabular}[c]{@{}c@{}}Execution Time\\ (milliseconds)\end{tabular} \\ \hline
rTPNN                  & 210   & 0.17   \\ \hline
LSTM                   & 70    & 0.14   \\ \hline
MLP                    & 47    & 0.08   \\ \hline
\begin{tabular}{c} Random \\ Forest \end{tabular}         & 11.8  & 0.12   \\ \hline
\begin{tabular}{c} Linear \\ Regression \end{tabular}      & 0.004 & 0.0025 \\ \hline
\begin{tabular}{c} Lasso \\ Regression \end{tabular}      & 0.005 & 0.0012 \\ \hline
\begin{tabular}{c} Ridge \\ Regression \end{tabular}  & 0.004 & 0.0012 \\ \hline
\begin{tabular}{c} Elastic Net \\ Regression \end{tabular} & 0.007 & 0.0012 \\ \hline
\end{tabular}\label{table:forecaster_times}
\end{table}

\begin{figure}[h!]
	\centering
	\includegraphics[scale=0.24]{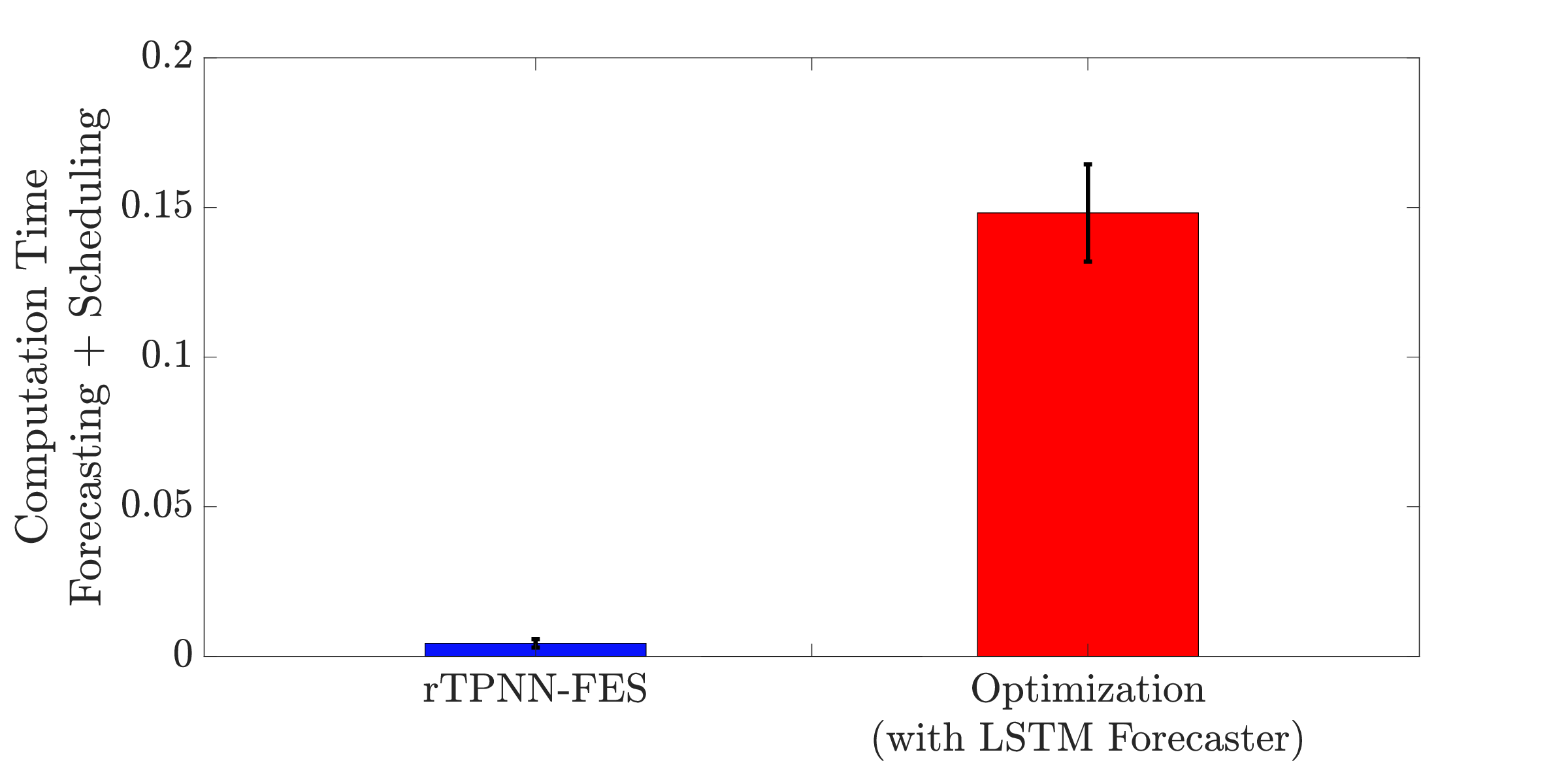}
	\caption{Computation time (in seconds) comparison between rTPNN-FES and optimal scheduling under LSTM forecaster}
	\label{fig:comp_time_sch}
\end{figure}

Figure~\ref{fig:comp_time_sch} displays the computation time of rTPNN-FES and that of optimization combined with LSTM (the second-best forecaster after rTPNN) in seconds. {Note that we do not present the computation of GA-based scheduling in this figure since it takes $4.61$ seconds on average -- which is approximately 3 orders of magnitude higher than the computation time of rTPNN-FES and 1 order of magnitude higher than that of optimization -- to find a schedule for a single window.} 
Our results in this figure show that rTPNN-FES requires significantly lower computation time than optimization to generate a daily schedule of household appliances. The average computation time of rTPNN-FES is about $4~ms$ while that of optimization with LSTM is $150~ms$. That is, rTPNN-FES is $37.5$ times faster than optimization with LSTM to simultaneously forecast and schedule. Although the absolute computation time difference seems insignificant for a small use case (as in this paper), it would have important effects on the operation of large renewable energy networks with a high number of sources and devices. 

\section{Conclusion}\label{sec:Conclusion}

We have proposed a novel neural network architecture, called Recurrent Trend Predictive Neural Network based Forecast Embedded Scheduling (namely rTPNN-FES), for smart home energy management systems. The rTPNN-FES architecture forecasts renewable energy generation and schedules household appliances to use renewable energy efficiently and to minimize user dissatisfaction. As the main contribution of rTPNN-FES, it performs both forecasting and scheduling in a single architecture. Thus, it 1) provides a schedule that is robust against forecasting and measurement errors, 2) requires significantly low computation time and memory space by eliminating the use of two separate algorithms for forecasting and scheduling, and 3) offers high scalability to grow the load set (i.e. adding devices) over time. 

We have evaluated the performance of rTPNN-FES for both forecasting renewable energy generation and scheduling household appliances using two publicly available datasets. During the performance evaluation, rTPNN-FES is compared against 8 different techniques for forecasting and against the optimization {and genetic algorithm }for scheduling. Our experimental results have drawn the following conclusions: 
\begin{itemize}
    \item  The forecasting layer of rTPNN-FES outperforms all of the other forecasters for the majority of MSE, MAE, MAPE, and SMAPE metrics.
    \item rTPNN-FES achieves {a highly} successful schedule which is very close to the optimal schedule with only $1.3\%$ of the cost difference.
    \item rTPNN-FES requires a much shorter time than both optimal and GA-based scheduling to generate embedded forecasts and scheduling, although the forecasting time alone is slightly higher than other forecasters.
\end{itemize}

Future work shall improve the training of rTPNN-FES by directly minimizing the cost of user dissatisfaction (or other scheduling costs) to eliminate the collection of optimal schedules for training. {In addition, the integration of a predictive dynamic thermal model into the rTPNN-FES framework shall be pursued in future studies. (Such integration is required to utilize more advanced HVAC scheduling/control system designs.)} It would also be interesting to observe the performance of rTPNN-FES for large-scale renewable energy networks. Furthermore, since the architecture of rTPNN-FES is not dependent on the particular optimization problem formulated in this paper, rTPNN-FES shall be applied for other forecasting/scheduling problems such as optimal dispatch in microgrids, flow control in networks, and smart energy distribution in future work.

\bibliographystyle{elsarticle-num}
\bibliography{references_techniques}

\end{document}